\documentclass[final]{cvpr}

\usepackage{times}
\usepackage{epsfig}
\usepackage{graphicx}
\usepackage{amsmath}
\usepackage{amssymb}
\usepackage{wrapfig}
\usepackage{multirow}
\usepackage{subcaption}
\usepackage{diagbox}

\newcommand\blfootnote[1]{%
  \begingroup
  \renewcommand\thefootnote{}\footnote{#1}%
  \addtocounter{footnote}{-1}%
  \endgroup
}

\usepackage{comment}
\usepackage{fixltx2e}
\usepackage[dvipsnames]{xcolor}

\usepackage[pagebackref=true,breaklinks=true,letterpaper=true,colorlinks,bookmarks=false]{hyperref}

\begin{document}

\title{3D CNNs with Adaptive Temporal Feature Resolutions}

\author{
Mohsen Fayyaz$^{1,*}$, Emad Bahrami$^{1,*}$, \\Ali Diba$^{2}$, Mehdi Noroozi$^{3}$, Ehsan Adeli$^{4}$, Luc Van Gool$^{2,5}$, Juergen Gall$^{1}$\\
$^{1}$University of Bonn, $^{2}$KU Leuven, $^{3}$Bosch Center for Artificial Intelligence,\\ $^{4}$Stanford University, $^{5}$ETH Z\"{u}rich \\
{\tt\small \{lastname\}@iai.uni-bonn.de, \{firstname.lastname\}@kuleuven.be, }\\
{\tt \small mehdi.noroozi@de.bosch.de, eadeli@stanford.edu}
}

\maketitle


\begin{abstract}

While state-of-the-art 3D Convolutional Neural Networks (CNN) achieve very good results on action recognition datasets, they are computationally very expensive and require many GFLOPs. While the GFLOPs of a 3D CNN can be decreased by reducing the temporal feature resolution within the network, there is no setting that is optimal for all input clips. In this work, we therefore introduce a differentiable Similarity Guided Sampling (SGS) module, which can be plugged into any existing 3D CNN architecture. SGS empowers 3D CNNs by learning the similarity of temporal features and grouping similar features together. As a result, the temporal feature resolution is not anymore static but it varies for each input video clip. By integrating SGS as an additional layer within current 3D CNNs, we can convert them into much more efficient 3D CNNs with adaptive temporal feature resolutions (ATFR). Our evaluations show that the proposed module improves the state-of-the-art by reducing the computational cost (GFLOPs) by half while preserving or even improving the accuracy. We evaluate our module by adding it to multiple state-of-the-art 3D CNNs on various datasets such as Kinetics-600, Kinetics-400, Mini-Kinetics, Something-Something~V2, UCF101, and HMDB51.

\end{abstract}
\blfootnote{$^{\star}$Mohsen Fayyaz and Emad Bahrami equally contributed to this work. Emad Bahrami contributed to this project while he was a visiting researcher at the Computer Vision Group of the University of Bonn.}

\section{Introduction}

\begin{figure*}[t]
    \centering \vspace{0pt}
    \includegraphics[width=1.0\linewidth]{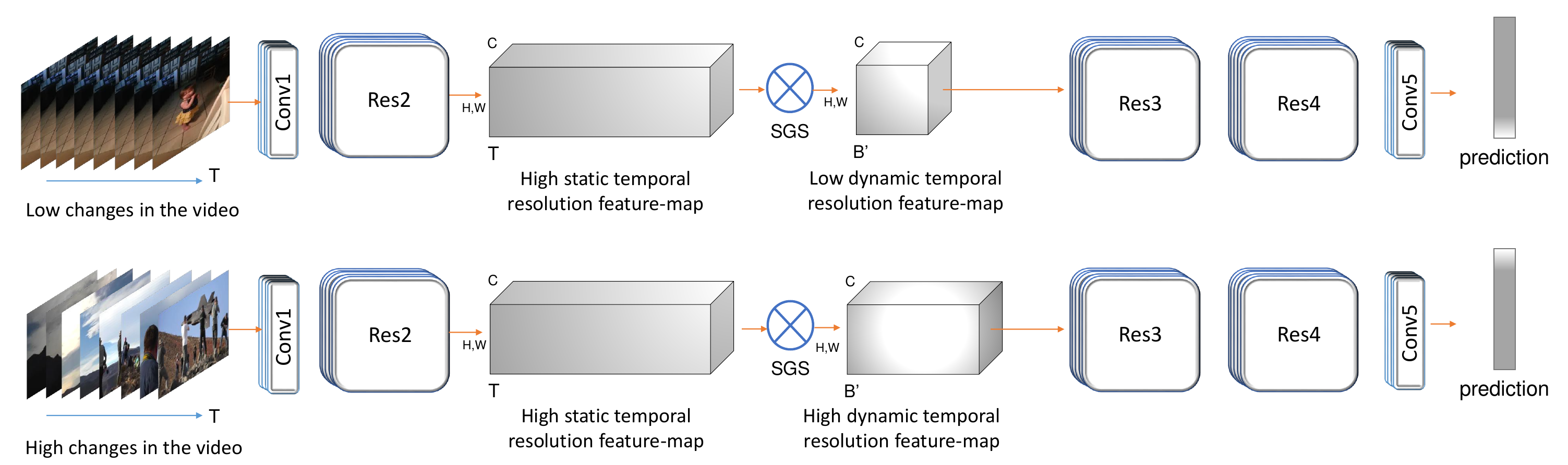} \vspace{0pt}
    \caption{The difficulty of recognizing actions varies largely across videos. For videos with slow motion (top), the temporal features that are processed within a 3D CNN can be highly redundant. However, there are also very challenging videos where all features are required to understand the content (bottom). While previous 3D CNNs use fix down-sampling schemes that are independent of the input video, we propose a similarity guided sampler that groups and aggregates redundant information of temporal features into $B' \leq T$ feature maps. The core aspect is that this process adapts the internal temporal resolution to the input video such that $B'$ is small if the input features are redundant (top) and large (bottom) if most of the features are required. } \vspace{0pt} 
    \label{fig:DTSN}
\end{figure*}


In recent years, there has been a tremendous progress for video processing in the light of new and complex deep learning architectures, which are based on variants of 3D Convolutional Neural Networks (CNNs)~\cite{c3d, slowfast, tle, stcnn,dynamonet, res3d, x3d}. They are trained for a specific number of input frames, typically between 16 to 64 frames. For classifying a longer video, they slide over the video and the outputs are then aggregated.
These networks, however, are often very expensive to train and heavy to deploy for inference task.  In order to reduce the inference time, \cite{SCSampler, FASTER} proposed to process not all parts of a video with the same 3D CNN. While~\cite{SCSampler} trains a second network that decides for each chunk of input frames if it should be processed by the more expensive 3D CNN, \cite{FASTER} uses a fix scheme where a subset of the input chunks are processed by an expensive 3D CNN and the other chunks by a less expensive 3D CNN. The latter then uses an RNN to fuse the outputs of the different 3D CNNs. Although both approaches effectively reduce the GFLOPS during inference, they increase the training time since two instead of one network need to be trained. Furthermore, they do not reduce the computational cost of the 3D CNNs themselves.

In this work, we propose an approach that makes 3D CNNs more efficient for training and inference. Our proposal is based on the observation that the computational cost of a 3D CNN depends on the temporal resolution it operates on at each stage of the network. While the temporal resolution can be different at different stages, the schemes that define how the temporal resolution is reduced is hard-coded and thus the same for all videos. However, it is impossible to define a scheme that is optimal for all videos. If the temporal resolution is too much reduced, the network is forced to discard important information for some videos. This results in a decrease of the action recognition accuracy performance. Vice versa, a high temporal resolution results in highly redundant feature maps and increases the computational time, which makes the 3D CNN highly inefficient for most videos. In this work, we therefore address the question of how a 3D CNN can dynamically adapt its computational resources in a way such that not more resources than necessary are used for each input chunk. 

In order to address this question, we propose to exploit the redundancy within temporal features such that 3D CNNs process and select the most valuable and informative temporal features for the action classification task. In contrast to previous works, we propose to dynamically adapt the temporal feature resolution within the network to the input frames such that on one hand important information is not discarded and on the other hand no computational resources are wasted for processing redundant information. 
To this end, we propose a \emph{Similarity Guided Sampling  (SGS)} mechanism that measures the similarity of temporal feature maps, groups similar feature maps together, and aggregates the grouped feature maps into a single output feature map. The similarity guided sampling is designed such that it is differentiable and number of output feature maps varies depending on the redundancy of the temporal input feature maps as shown in Fig.~\ref{fig:DTSN}. By integrating the similarity guided sampling as an additional module within any 3D CNN, we convert the 3D CNN with fixed temporal feature resolutions into a much more efficient dynamic 3D CNN with \emph{adaptive temporal feature resolutions (ATFR)}. Note that this approach is complementary to \cite{SCSampler, FASTER} and the two static 3D CNNs used in these works can be replaced by adaptive 3D CNNs. However, even with just a single 3D CNN with adaptive temporal feature resolutions, we already achieve a higher accuracy and lower GFLOPs performance compared to \cite{SCSampler, FASTER}.  

We demonstrate the efficiency of 3D CNNs with adaptive temporal feature resolutions by integrating the similarity guided sampler into the current state-of-the-art 3D CNNs such as R(2+1)D~\cite{R21}, I3D~\cite{i3d}, and X3D~\cite{x3d}. It drastically decreases the GFLOPs by about half in average while the accuracy remains nearly the same or gain improvements. 
In summary, the similarity guided sampler is capable of significantly scaling down the computational cost of off-the-shelf 3D CNNs and therefore plays a crucial role for real-world video-based applications.

\section{Related Work}

The computer vision community has made huge progress in several challenging vision tasks by using CNNs. In recent years, there has been a tremendous progress for video processing in the light of new and complex deep learning architectures, which are based on variants of 3D CNNs \cite{c3d, slowfast, tle, stcnn,dynamonet, res3d, x3d}. Tran et al.~\cite{c3d} and Carreira et al.~\cite{i3d} proposed 3D versions of VGG and Inception architectures for large-scale action recognition benchmarks like Sports-1M~\cite{sports1m} and Kinetics~\cite{kinetics400}. These methods could achieve superior performance even without using optical-flow or any other pre-extracted motion information. This is due to the capability of 3D kernels to extract temporal relations between sequential frames. Recently, methods like HATNet \cite{hatnet}, STC~\cite{stcnn}, and DynamoNet \cite{dynamonet} focus on exploiting spatial-temporal correlations in a more efficient way or on learning more accurate motion representations for videos. 
These works based on 3D CNNs, however, require huge computational resources since they process sequences of frames with an immense number of 3D convolution layers.
There has been therefore a good effort to propose more efficient architectures based on 2D and 3D CNNs~\cite{tsm, 27TSM,R21,53TSM, 61tsm}. For instance, Lin et al.~\cite{tsm} introduced a temporal shift module (TSM) to enhance 2D-ResNet CNNs for video classification. The model even runs on edge devices. 
In~\cite{R21,53TSM} 2D and 3D convolutional layers are combined in different ways. SlowFast \cite{slowfast} has explored the resolution trade-offs across temporal, spatial, and channel axes. It decreases the computation cost by employing a light pathway with a high temporal resolution for temporal information modeling and a heavy low temporal resolution pathway for spatial information modeling. In relation to this work, \cite{x3d} investigates whether the light or heavy model is required and presents X3D as a family of efficient video networks.

In order to reduce the inference time of existing networks, \cite{liteeval, adaframe, SCSampler, FASTER} proposed to process not all parts of a video with the same CNN model. This line of research is built upon the idea of big-little architecture design. In the context of 2D CNNs, \cite{liteeval, adaframe} process salient frames by expensive models and use light models to process the other frames. In contrast, \cite{SCSampler,FASTER} do not process single frames but process short chunks of frames with 3D CNNs. \cite{SCSampler} trains a second lighter network that decides for each chunk of input frames if it should be processed by the more expensive 3D CNN. \cite{FASTER} uses a fix scheme where a subset of the input chunks are processed by an expensive 3D CNN and the other chunks by a less expensive 3D CNN. It then uses an RNN to fuse the outputs of the different 3D CNNs. Although such approaches effectively reduce the GFLOPS during inference, they increase the training time since two instead of one network need to be trained. Furthermore, they do not reduce the computational cost of the 3D CNNs themselves.

There are various efforts on temporal action detection or finding action segments in untrimmed videos like~\cite{Actionsearch,stones,actglimpse,Multiagent}. These works focus on localizing actions but not on improving the computational efficiency for action recognition. While these works are not related, they can benefit from our approach by integrating the proposed similarity guided sampling module into their CNNs for temporal action localization.

\section{Adaptive Temporal Feature Resolutions}

Current state-of-the-art 3D CNNs operate at a static temporal resolution at all levels of the network. Due to the redundancy of neighbouring frames, traditional 3D CNN methods often down-sample the temporal resolution inside the network. This helps the model to operate at a lower temporal resolution and hence reduces the computation cost. The down-sampling, however, is static which has disadvantages in two ways. First, a fixed down-sampling rate can discard important information, in particular for videos with very fast motion as it is for instance the case for ice hockey games. Second, a fixed down-sampling rate might still include many redundant temporal features that do not contribute to the classification accuracy as it is for instance the case for a video showing a stretching exercise. We therefore propose a module that dynamically adapts the temporal feature resolution within the network to the input video such that on one hand important information is not discarded and on the other hand no computational resources are wasted for processing redundant information.

\begin{figure}
\centering
    \vspace{0pt}
    \includegraphics[width=1.0\columnwidth]{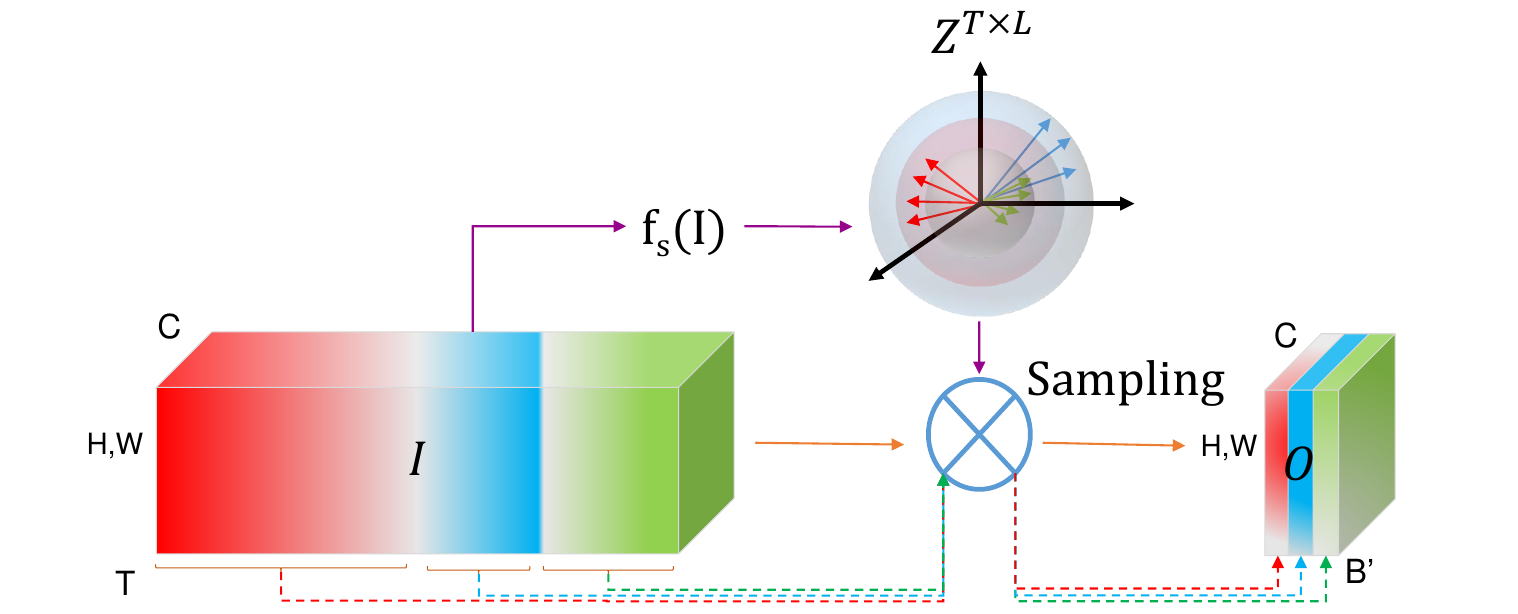} \vspace{0pt}
    \caption{To learn the similarity of the feature maps, we map each temporal feature map $\mathcal{I}_t$ using $f_s(\mathcal{I})$ into an $L$-dimensional similarity space. After the mapping, $\mathcal{Z} \in \mathbb{R}^{T\times L}$ contains all of the feature maps represented as vectors in the similarity space. Afterwards, we group similar vectors by creating B similarity bins. Using the similarity bins, sampler aggregates the similar feature maps of each bin into the output feature map $\mathcal{O}_b$.}
    \label{fig:SGS}\vspace{0pt}
\end{figure}



Fig.~\ref{fig:DTSN} illustrates a 3D CNN with \emph{adaptive temporal feature resolutions (ATFR)}. The core aspect of ATFR is to fuse redundant information from a temporal sequence of features and extract only the most relevant information in order to reduce the computational cost for processing a video. An important aspect is that this approach is not static, \ie, the amount of information that is extracted varies for each video as illustrated in Fig.~\ref{fig:DTSN}. In order to achieve this, we propose a novel \emph{Similarity Guided Sampling (SGS)}  mechanism that will be described in Sec.~\ref{sec:similarityGuidedSampler}. 


In principle, any 3D CNN can be converted into a CNN with adaptive temporal feature resolutions by using our SGS module. Since the module is designed to control the temporal resolution within the network for each video, it should be added to the early stages of a network in order to get the best reduction of computational cost. For R(2+1)D~\cite{R21}, for example, we recommend to add SGS after the second ResNet block. This means that the temporal resolution is constant for all videos before SGS, but it dynamically changes after SGS. We discuss different 3D CNNs with SGS in Sec.~\ref{sec:ATFRImp}.

\section{Similarity Guided Sampling}
\label{sec:similarityGuidedSampler}

The \emph{SGS} is a differentiable module to sample spatially similar feature maps over the temporal dimension and aggregate them into one feature map. Since the number of output feature maps is usually lower than the input feature maps, i.e., $B' < T$, redundant information is removed as illustrated in Figure \ref{fig:SGS}. The important aspect is that $B'$ is not constant, but it varies for each video. In this way, we do not remove any information if there is no redundancy among the input feature maps.  

This means that we need to a) learn the similarity of feature maps, b) group similar feature maps, and c) aggregate the grouped feature maps. Furthermore, all these operations need to be differentiable. We denote an input feature map for frame $t$ by $\mathcal{I}_t \in \mathbb{R}^{C\times H \times W}$, where $C$, $H$, and $W$ denote the number of channels, height, and width, respectively.
To learn the similarity of the feature maps, we map each feature map $\mathcal{I}_t$ into an $L$-dimensional similarity space. This mapping $f_s(\mathcal{I}_t)$ is described in Sec.~\ref{sec:SimilaritySpace}. 
After the mapping, $\mathcal{Z} \in \mathbb{R}^{T\times L}$ contains all feature maps in the similarity space, which are then grouped and aggregated into $B'$ feature maps. The grouping of $\mathcal{Z}_t$ is described in Sec.~\ref{sec:SimilarityBins} and the aggregation of the grouped features in Sec.~\ref{sec:DifferentiableBinsSampling}.



\subsection{Similarity Space}
\label{sec:SimilaritySpace}
The similarity space is a $L$ dimensional vector space where each temporal input feature map is represented by a vector $\mathcal{Z}_t$.
The mapping is performed by the similarity network $f_s(\mathcal{I})$ that consists of a global average pooling layer and two convolutional layers. The pooling is applied over the spatial dimension of the feature map while keeping the temporal dimension. Afterward two $1D$ convolutional layers are applied with kernel sizes of $1$ and output channel sizes $C$ and $L$, respectively.

\subsection{Similarity Bins}
\label{sec:SimilarityBins}
To group similar feature maps $\mathcal{I}_t$, we use the magnitude of each vector $\mathcal{Z}_t$, i.e.,
\begin{equation}
    \label{eq:GeneralDistanceFunction}
    \Delta_{t} = ||\mathcal{Z}_t||
\end{equation}
and we consider two feature maps $\mathcal{I}_t$ and $\mathcal{I}_{t'}$ similar if the value of $\Delta_{t}$ and $\Delta_{t'}$ lie  inside a similarity bin. 
To make the grouping very efficient and differentiable, we propose a binning approach with $B$ similarity bins. We set $B=T$ such that no information is discarded if there is no redundancy between the feature maps of all frames. For most videos, a subset of bins remain empty and will be discarded such that the remaining bins, $B'$, will be less than $B$ as it is described in Sec. ~\ref{sec:DifferentiableBinsSampling}.


We first estimate the half of the width of each similarity bin $\gamma$, by computing the maximum magnitude $\Delta_{max}$ and dividing it by the number of the desired bins $B$:
\begin{equation}
    \label{eq:SimilarityBins}
    \Delta_{max}=\max(\Delta_{1},\dots,\Delta_{T}), \:\:
    \gamma=\frac{\Delta_{max}}{2B}.
\end{equation}
Having the width of the similarity bins, the center of each bin $\beta_b$ is estimated as follows:
\begin{equation}
    \label{eq:SimilarityBinsMean}
    \beta_{b}=(2b-1)\gamma \quad \forall b \in (1,\dots,B).
\end{equation}





\subsection{Differentiable Bins Sampling}
\label{sec:DifferentiableBinsSampling}
The grouping and aggregation of all feature maps $\mathcal{I}_t$ based on the bins $B$ will be done jointly by sampling temporal feature maps which belong to the same similarity bin and add them together. We denote the aggregated feature maps for each bin $b$ by $\mathcal{O}_b \in \mathbb{R}^{C\times H \times W}$. To make the process differentiable, we use generic differentiable sampling kernels $\Psi(.,\beta_b)$ that are defined such that a sampler only samples from the input temporal feature map $\mathcal{I}_t$ if $\Delta_t$ lies in the similarity bin $b$. This can be written as:
\begin{equation}
    \label{eq:GeneralSamplingKernel}
    \mathcal{O}_{b}=\sum_{t=1}^{T} \mathcal{I}_{t} \Psi(\Delta_{t}, \beta_{b}).
\end{equation}
Theoretically, any differentiable sampling kernel that has defined gradients or sub-gradients with respect to $\Delta_t$ can be used. In our experiments, we evaluate two sampling kernels. The first kernel is based on  
the Kronecker-Delta function $\delta$:
\begin{equation}
    \label{eq:DeltaSamplingKernel}
    \mathcal{O}_{b}=\frac{1}{\sum_{t=1}^{T} \delta\left(\left\lfloor \frac{|\Delta_{t} - \beta_{b}|}{\gamma} \right\rfloor\right)}\sum_{t=1}^{T} \mathcal{I}_{t} \delta\left(\left\lfloor \frac{|\Delta_{t} - \beta_{b}|}{\gamma} \right\rfloor\right).
\end{equation}
The kernel averages the feature maps that end in the same bin. As second kernel, we use a linear sampling kernel:
\begin{equation}
    \label{eq:LinearSamplingKernel}
    \mathcal{O}_{b}=\sum_{t=1}^{T} \mathcal{I}_{t} \max\left(0, 1 - \frac{|\Delta_{t} - \beta_{b}|}{\gamma}\right) .
\end{equation}
The kernel gives a higher weight to feature maps that are closer to $\beta_{b}$ and less weights to feature maps that are at the boundary of a bin. While we evaluate both kernels, we use the linear kernel by default.   

After the sampling, some bins remain empty, i.e., \mbox{$\mathcal{O}_b=0$}. 
We drop the empty bins and denote by $B'$ the bins that remain. Note that $B'$ varies for each video as illustrated in Fig.~\ref{fig:DTSN}. In our experiments we show that the similarity guided sampling can reduce the GFLOPS of a 3D CNN by over 47\% in average, making 3D CNNs suitable for applications where they are computationally expensive.     


\subsection{Backpropagation}
\label{sec:backPropagation}
Using differentiable kernels for sampling, gradients can be backpropagated through both $\mathcal{O}$ and $\Delta$, where $\Delta$ is the magnitude of the similarity vectors $\mathcal{Z}$ which are the outputs of $f_s(.)$. Therefore, we can backpropagate through $f_s(.)$. For the linear kernel \eqref{eq:LinearSamplingKernel}, which we use if not otherwise specified, the gradient with respect to $\mathcal{I}_t$ is given by
\begin{equation}
    \label{eq:differentiable_sampling_linear_I}
    \frac{\partial \mathcal{O}_b}{\partial \mathcal{I}_t}=\max\left(0, 1 - \frac{|\Delta_{t} - \beta_{b}|}{\gamma}\right)
\end{equation}
and the gradient with respect to $\Delta_t$ is given by
\begin{equation}
    \label{eq:differentiable_sampling_linear}
    \frac{\partial \mathcal{O}_b}{\partial \Delta_t}=\mathcal{I}_t \: 
    \begin{cases} 
    0 & |\beta_b - \Delta_t|\geq\gamma \\
    \frac{1}{\gamma} & \beta_b-\gamma < \Delta_t \leq \beta_b \\
    -\frac{1}{\gamma} & \beta_b < \Delta_t < \beta_b+\gamma
    \end{cases}.
\end{equation}
Note that for computing the sub-gradients \eqref{eq:differentiable_sampling_linear} only the kernel support region for each output bin needs to be considered. The sampling mechanism can therefore be efficiently implemented on GPUs.

\section{Experiments}

We evaluate our proposed method on the action recognition benchmarks Mini-Kinetics \cite{miniKinetics1}, Kinetics-400 \cite{kinetics400}, Kinetics-600 \cite{kinetics600}, Something-Something-V2 \cite{ssv}, UCF-101 \cite{ucfdataset}, and HMDB-51 \cite{hmdbdataset}. For these datasets, we use the standard training/testing splits and protocols provided by the datasets. For more details and the UCF-101 and HMDB-51 results please refer to the supplementary material.

\subsection{Implementation Details}
\label{sec:implementationDetails}

\textbf{3D CNNs with ATFR.}
\label{sec:ATFRImp}
The similarity guided sampling (SGS) is a differentiable module that can be easily implemented in current deep learning frameworks. We have implemented our SGS module as a new layer in PyTorch which can be easily added to any 3D CNN architecture. To better evaluate the SGS, we have added it to various backbones, such as R(2+1)D \cite{R21}, I3D \cite{i3d}, X3D \cite{x3d}, and a modified 3DResNet. We place our SGS layer on the second stage of the backbone models. Please refer to the supplementary material for more details. For all of the X3D based models, we follow the training, testing, and measurement setting in \cite{x3d} unless mentioned otherwise. Additional details and code are available online.\footnote{\url{https://SimilarityGuidedSampling.github.io}}

\textbf{Training.} 
Our models on Mini-Kinetics, Kinetics-400, and Kinetics-600 are trained from scratch using randomly initialized
weights without any pre-training. However, we fine-tune on Something-Something-V2, UCF-101, and HMDB-51 with models pre-trained on Kinetics-400. We trained our models using SGD with momentum 0.9 and a weight decay of 0.0001 following the setting in \cite{slowfast}. For Kinetics and Mini-Kinetics, we use a half-period cosine schedule \cite{halfCosineScheduler} with a linear warm-up strategy \cite{linearWarmUp} to adapt the learning rate over 196 epochs of training. During training, we randomly sample $32$ frames from a video with input stride $2$. For spatial transformations, we first scale the shorter side of each frame with a random integer from the interval between 256 and 320 \cite{non-local, slowfast, vgg} then we apply a random cropping with size $224 \times 224$ to each frame. Furthermore, each frame is horizontally flipped with probability of $0.5$. 

\textbf{Testing.}
We follow \cite{non-local, slowfast} and uniformly sample 10 clips from each video for inference. The shorter side of each clip is resized to $256$ and we extract 3 random crops of size $256 \times 256$ from each clip. For the final prediction, we average the softmax scores of all clips. 

\begin{figure}[]
    \centering\vspace{0pt}
    \includegraphics[width=0.75\columnwidth]{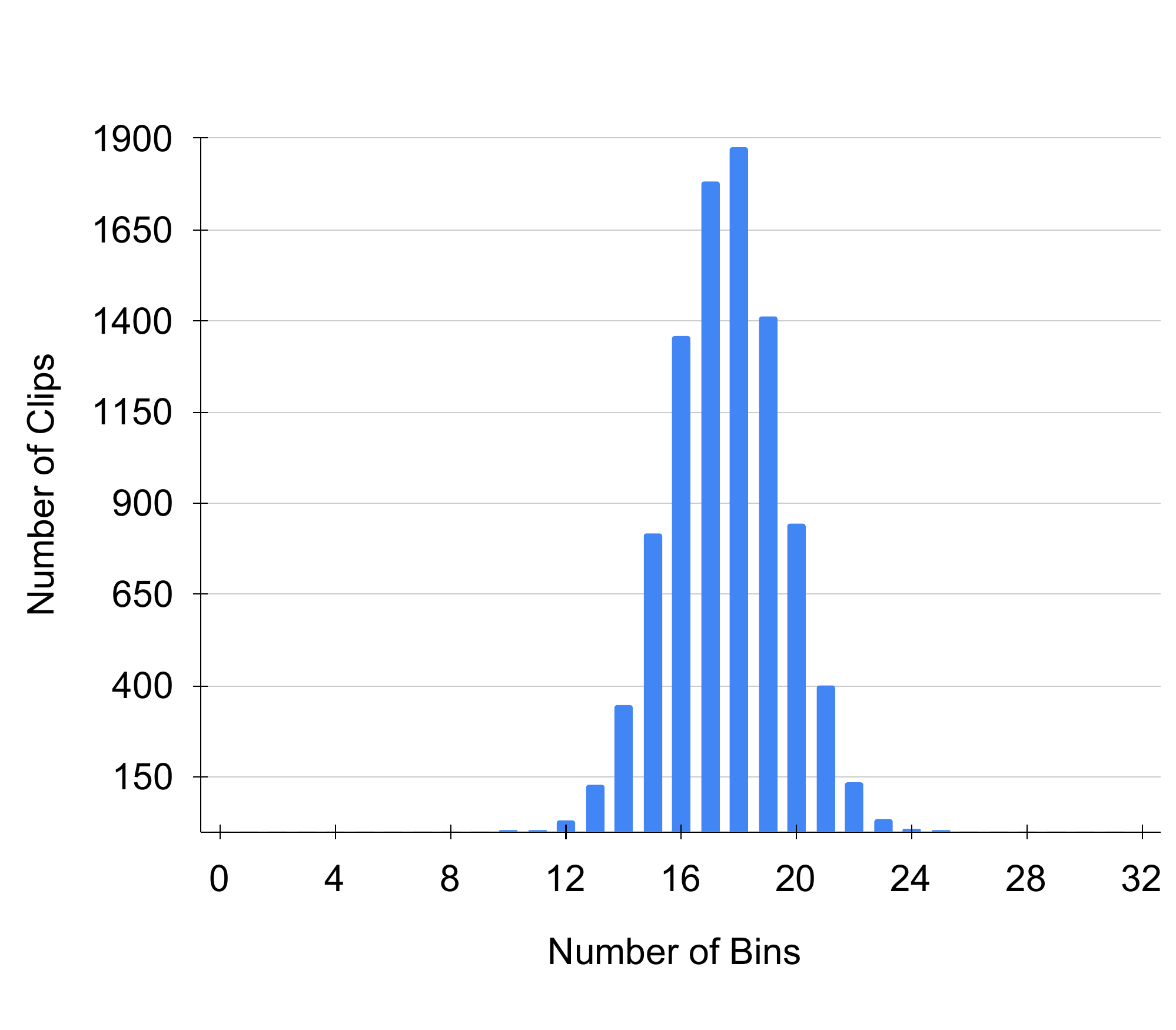}
    \caption{Histogram of active bins for 3DResNet-50 + ATFR on the Mini-Kinetics validation set. The y-axis corresponds to the number of clips and the x-axis to the number of active bins $B'$. } 
    \vspace{-15pt}
    \label{fig:hist_bins}
\end{figure}

\textbf{Measurements.}
We report top-1 and top-5 accuracy. To measure the computational efficiency of our models, we report the complexity of the models in GFLOPS based on a single input video sequence of 32 frames and spatial size 224 $\times$ 224 for validation and 256 $\times$ 256 for testing. As shown in Fig.~\ref{fig:hist_bins}, 3D CNNs with ATFR adapt the temporal feature resolutions and the GFLOPs vary for different clips. For ATFR models, we therefore report the average GFLOPs. 

\subsection{Ablation Experiments}
We first analyze different setups for our SGS module. Then, we analyze the efficiency and effect of using our SGS module in different 3D CNN models. If not otherwise specified, we use 3DResNet-18 as 3D CNNs backbones and report the results on the Mini-Kinetics validation set.

\subsubsection{Different Similarity Measurements}
As mentioned in Sec.~\ref{sec:SimilarityBins}, we use the magnitude of the embedding vectors as the similarity measurement to create the similarity bins. The embedding vectors are represented in an $L$ dimensional space. Instead of magnitudes, we can use other measures such as directions of the vectors. To better study this, we convert the Cartesian coordinates of the vectors to spherical coordinates. In an $L$ dimensional space, a vector is represented by $1$ radial coordinate and $L-1$ angular coordinates. To use the spherical coordinates of the vectors for creating the similarity bins, we use multi-dimensional bins and sampling kernels. For more details, please refer to the supplementary material. 

We report the results in Table \ref{table:similarityMeasurements}. As can be seen, using the magnitudes of the vectors results in a better accuracy compared to angular coordinates or spherical coordinates. We believe that due to the similarity of the neighbouring video frames using only magnitudes of the vectors for the similarity measurement is enough and angular or spherical coordinates add too much of complexity to the model. In all of the experiments, the number of bins $B$ is equal to $32$. For the angular coordinates, we divide the angles into $4$ and $8$ bins $(4\times8)$. For the spherical coordinates, we divide the radial coordinate into $2$ and the angular coordinates into $4$ and $4$ bins $(2\times4\times4)$.

\begin{table}[t]
    \centering
    \begin{tabular}{ |c| c c c |}
      \hline
      Similarity & Magnitude & Angular & Spherical\\
      \hline 
      top1 & \textbf{69.6} & 68.5 & 68.7 \\
      top5 & \textbf{88.8} & 87.8 & 88.1 \\ \hline
    \end{tabular} \hfill
    \caption{Impact of the similarity measure for 3DResNet-18 + ATFR on Mini-Kinetics with linear sampling kernel. We show top-1 and top-5 classification accuracy (\%).}
    \vspace{-8pt}
    \label{table:similarityMeasurements}
\end{table}

\subsubsection{Different Sampling Kernels}
As mentioned in Sec.~\ref{sec:DifferentiableBinsSampling}, we can use different differentiable sampling kernels (\ref{eq:GeneralSamplingKernel}). We evaluate two different sampling kernels, namely the Kronecker-Delta sampling kernel (\ref{eq:DeltaSamplingKernel}) and the linear sampling kernel (\ref{eq:LinearSamplingKernel}). As can be seen in Table \ref{tab:samplingKernels}, the linear kernel performs better than the Kronecker-Delta kernel. The slight superiority of the linear kernel is due to the higher weighting of the temporal feature maps that are closer to the center of the bins. We use the linear kernel for the rest of the paper. 

\begin{table}[t]
    \centering
    \begin{tabular}{ |c| c c|}
    \hline
    Kernel  & Linear & Kronecker \\
    \hline 
    top1 & \textbf{69.6} & 68.9 \\
    top5 & \textbf{88.8} & 88.6 \\
    \hline
    \end{tabular}
    \caption{Impact of the sampling kernel.
    }
    \vspace{-10px}
    \label{tab:samplingKernels}
\end{table}

 


\subsubsection{Embedding Dimension}

As mentioned in Sec.~\ref{sec:SimilaritySpace}, we map the temporal feature maps into an $L$-dimensional similarity space. In Table \ref{table:embeddingDimension}, we quantify the effect of $L$. The accuracy increases as $L$ increases until $L=8$. For $L=16$ the dimensionality is too large and the similarity space tends to overfit. The model with $L=1$ is a special case since it can be considered as a direct prediction of $\Delta_t$  \eqref{eq:GeneralDistanceFunction} without mapping the temporal features into a similarity space $\mathcal{Z}_t$.
The results show that using a one dimensional embedding space results in a lower accuracy, which demonstrates the benefit of the similarity space.

\begin{table}[t] 
\centering
\small
\tabcolsep=0.15cm

\begin{tabular}{ |c| c c c c|}
\hline
  $L$  & 1 & 4 & 8 & 16 \\
  \hline 
  top1 & 67.3  & 68.4 & \textbf{69.6} & 64.7 \\
  top5 & 87.7  & 88.1 & \textbf{88.8} & 86.1 \\
  \hline
 
\end{tabular}
\caption{Impact of the dimensionality $L$ of the similarity space.}
\vspace{-5pt}
\label{table:embeddingDimension}
\end{table}

\subsubsection{Different Input Frame-rates} It is an interesting question to ask how a 3D CNN with ATFR performs when the number of input frames or the stride changes for inference. To answer this question, we train two 3D CNNs with ATFR and two without ATFR using 32 input frames and a sampling stride of 2, which corresponds to a temporal receptive field of 64 frames. For inference, we then change the number of frames to 64 and/or the stride to 1.         

As it can be seen in Table~\ref{tab:inputFrameRate}, increasing the input frames from 32 to 64 improves the accuracy of all models. This improvement in accuracy is due to the increase in the temporal receptive field over the input frames while keeping the temporal input resolution.
However, the computation cost of the models without ATFR increases as expected by factor 2. If ATFR is used, the increase is only by 1.7 and 1.5 for R(2+1)D+ATFR and 3DResNet-18+ATFR. By comparing R(2+1)D with R(2+1)D+ATFR, we see how ATFR drastically reduces the GFLOPS from 46.5 to 32.3 for 32 frames and from 93.1 to 54.9 for 64 frames. This shows that more frames also increase the redundancy and ATFR efficiently discards this redundancy. Furthermore, it demonstrates that ATFR is robust to changes of the frame-rate and number of input frames. 

It is also interesting to compare the results for 32 frames with stride 2 to the results for 64 frames with stride 1. In both cases, the temporal receptive field is 64. We can see the efficiency of our method in adapting the temporal resolution compared to the traditional static frame-rate sampling methods, \ie, 3DResNet-18+ATFR operates on average with 21.1 GFLOPs for 64 input frames compared to SlowFast with GFLOPs of 30.9 (32) and 61.8 (64), and R(2+1)D with GFLOPs of 46.5 (32) and 93.1 (64). 

\begin{table}[t] 
\centering
\small
\tabcolsep=0.25cm
\resizebox{1.0\columnwidth}{!}{
\begin{tabular}{|c|c|c|clclclcl|}
\hline
\multirow{3}{*}{model} & \multirow{3}{*}{input frames} & \multirow{3}{*}{GFLOPs}  & \multicolumn{4}{c|}{top1}                                & \multicolumn{4}{c|}{top5} \\ \cline{4-11}
&                              &                         &              \multicolumn{8}{c|}{stride}  \\ \cline{4-11}
                              &                              &                         &              \multicolumn{2}{c|}{1} & \multicolumn{2}{c|}{2} & \multicolumn{2}{c|}{1} & \multicolumn{2}{c|}{2} \\ \hline

\multirow{2}{*}{SlowFast-8x8-ResNet18}  & \multirow{1}{*}{32}           & \multirow{1}{*}{30.9}                             & \multicolumn{2}{c|}{67.5}  & \multicolumn{2}{c|}{69.7}   & \multicolumn{2}{c|}{87.1}  & \multicolumn{2}{c|}{89.1}  \\ \cline{2-11} 
                              
                              & \multirow{1}{*}{64}           & \multirow{1}{*}{61.8 (2.0)}                    & \multicolumn{2}{c|}{72.1}  & \multicolumn{2}{c|}{74.6}  & \multicolumn{2}{c|}{89.9}  & \multicolumn{2}{c|}{91.9}  \\
                              \hline \hline
\multirow{2}{*}{R(2+1)D} & \multirow{1}{*}{32}           & \multirow{1}{*}{46.5}                             & \multicolumn{2}{c|}{67.4}  & \multicolumn{2}{c|}{69.3}  & \multicolumn{2}{c|}{86.2}  & \multicolumn{2}{c|}{87.5}  \\ \cline{2-11} 
                              
                              & \multirow{1}{*}{64}           & \multirow{1}{*}{93.1 (2.0)}                             & \multicolumn{2}{c|}{70.8}  & \multicolumn{2}{c|}{73.7}  & \multicolumn{2}{c|}{88.8}  & \multicolumn{2}{c|}{91.5}  \\
                              \hline \hline
\multirow{2}{*}{R(2+1)D+ATFR} & \multirow{1}{*}{32}           & \multirow{1}{*}{$32.3$}                             & \multicolumn{2}{c|}{67.4}  & \multicolumn{2}{c|}{69.3}  & \multicolumn{2}{c|}{86.4}  & \multicolumn{2}{c|}{87.6}  \\ \cline{2-11} 
                              
                              & \multirow{1}{*}{64}           & \multirow{1}{*}{$54.9$ (1.7)}       &      \multicolumn{2}{c|}{71.4}  & \multicolumn{2}{c|}{73.8}  & \multicolumn{2}{c|}{88.6}  & \multicolumn{2}{c|}{90.7}  \\
                              \hline \hline

\multirow{2}{*}{3DResNet-18+ATFR}     & \multirow{1}{*}{32}           & \multirow{1}{*}{\textbf{14.0}}        & \multicolumn{2}{c|}{67.3}  & \multicolumn{2}{c|}{69.6}  & \multicolumn{2}{c|}{87.1}  & \multicolumn{2}{c|}{88.8}  \\ 
                              \cline{2-11} 
                              & \multirow{1}{*}{64}           & \multirow{1}{*}{\textbf{21.1 (1.5)} }               & \multicolumn{2}{c|}{72.1}  & \multicolumn{2}{c|}{74.8}  & \multicolumn{2}{c|}{89.8}  & \multicolumn{2}{c|}{91.4} \\ \hline
                              
\end{tabular}
}
\caption{Impact of the stride and number of input frames during inference. All models are trained with 32 frames and stride 2. 
} 
\vspace{-15pt}
\label{tab:inputFrameRate}
\end{table}
\subsubsection{Adaptive Temporal Feature Resolutions}
As shown in Fig.~\ref{fig:hist_bins}, the temporal feature resolutions vary for different clips. In order to analyze how the temporal feature resolution relates to the content of a video, we report in Table \ref{tab:action_atfr} the 5 action classes with lowest adaptive temporal feature resolution ($<$12) and highest adaptive temporal feature resolution ($>$20). As in Fig.~\ref{fig:hist_bins}, the results are for the 3DResNet-50+ATFR on the Mini-Kinetics validation set.
As it can be seen, the actions with less movements like `presenting weather forecast' result in a low temporal resolution while actions with fast (camera) motions like `passing American football (in game)' result in a high temporal resolution.
\begin{table}[t]
    \centering
    \resizebox{1.0\columnwidth}{!}{
    \begin{tabular}{ |l| r| }
    \hline
    Lowest Temporal Resolution & Highest Temporal Resolution \\
    \hline 
     presenting weather forecast & passing American football (in game)  \\
     stretching leg & swimming breast stroke \\
     playing didgeridoo & playing ice hockey \\
     playing clarinet & pushing cart \\
     golf putting & gymnastics tumbling\\ \hline
    \end{tabular}
    }
    \caption{The 5 action classes with lowest and highest required adaptive temporal resolution for 3DResNet-50 + ATFR on Mini-Kinetics.}
    \vspace{-8pt}
    \label{tab:action_atfr}
\end{table}

\subsubsection{SGS Placement}
To evaluate the effect of the location of our SGS module within a 3D CNN, we add it to different stages of X3D-S \cite{x3d} and train it on Mini-Kinetics. As it can be seen in Table \ref{tab:ATFRLevel}, adding SGS to the first stage of X3D-S drastically reduces the GFLOPs by {52.6\% (2.1$\times$)} while getting slightly lower accuracy. On the other hand, adding SGS after the 2$^{nd}$ stage results in a 42.1\% reduction of GFLOPs and slightly higher accuracy. The same accuracy and growth in GFLOPs occurs when SGS is added after the 3$^{rd}$ stage. 
\begin{table}[t]
    \centering
    \resizebox{1.0\columnwidth}{!}{
    \begin{tabular}{ |c| c c c c|}
    \hline
    Stage  & No SGS & First Conv & Res2 & Res3 \\
    \hline 
    top1 & 77.9 &77.8 & \textbf{78.0} & 78.0 \\
    GFLOPs & 1.9 & \textbf{0.9} & 1.1 & 1.3 \\ \hline
    \end{tabular}}
    \caption{Evaluating the result of adding our SGS layer to different stages of a X3D-S network on Mini-Kinetics.}
    \vspace{-12pt}
    \label{tab:ATFRLevel}
\end{table}

\subsection{Mini-Kinetics}
Mini-Kinetics is a smaller dataset compared to the full Kinetics-400 dataset \cite{kinetics400} and consists of 200 categories. Since some videos on YouTube are not accessible, the training and validation set contain 144,132 and 9182 video clips, respectively. Table~\ref{tab:minikinetics} shows the results on Mini-Kinetics. We add the SGS module to four 3d CNNs R(2+1)D \cite{R21}, I3D \cite{i3d}, X3D \cite{x3d}, and 3DResNet. In all cases, ATFR drastically reduces the GFLOPS while the accuracy remains nearly the same. For X3D, the accuracy even increases marginally.

\begin{table}[t]
    \centering
    \resizebox{1.0\columnwidth}{!}{
    \begin{tabular}{|l|c|c|c|c|}
    \hline
        \multicolumn{1}{|c|}{model} & \multirow{1}{*}{backbone} & \multirow{1}{*}{GFLOPs} &  \multicolumn{1}{c|}{top1}  & top5  \\ \hline
        Fast-S3D \cite{miniKinetics1}                & -                  &      43.5               &  78.0                       & - \\
        \hline
        SlowFast 8x8           & ResNet18                  & 40.4                    & 77.5                       & 93.3  \\ \hline
        SlowFast 8x8           & ResNet50                  & 65.7                    & 79.3                       & 94.2  \\ \hline \hline
        R(2+1)D                & ResNet50                  &      101.8               & 78.7                       & 93.4  \\ 
        R(2+1)D\textbf{+ATFR}                & ResNet50                  &      67.3               & 78.2                       & 92.9  \\
        \hline
        I3D                & ResNet50                  &      148.4               & 79.3                       & 94.4  \\ 
        I3D\textbf{+ATFR}                & ResNet50                  &      105.2               &  78.8                       & 93.6 \\
        \hline
        X3D-S                   & -                  & 1.9                  & 77.9                       & 93.4 \\
        X3D-S\textbf{+ATFR}     & -                  & \textbf{1.1}         & 78.0                       & 93.5 \\ \hline
        3DResNet                   & ResNet50                  &        40.8           &        79.2                &  94.6 \\ 
        3DResNet\textbf{+ATFR}                   & ResNet50                  & \textbf{23.4}                    & 79.3                       & 94.6 \\  \hline
        \end{tabular}} \hfill
    \caption{Comparison with state-of-the-art methods on Mini-Kinetics. The accuracy for Fast-S3D \cite{miniKinetics1} is reported with 64 frames. }
    \vspace{-15pt}
    \label{tab:minikinetics}
\end{table}

\subsection{Kinetics-400}
We also evaluate ATFR with state-of-the-art 3D CNNs on Kinetics-400 \cite{kinetics400}, which contains $\sim$240k training and $\sim$20k validation videos of 400 human action categories. Table \ref{tab:Kinetics400} shows the comparison with the state-of-the-art. We add the SGS module to the state-of-the-art 3D CNNs SlowFast \cite{slowfast} and three versions of X3D \cite{x3d}.

As it can be seen, our {SGS} module drastically decreases the GFLOPs of all 3D CNNs. In contrast to Mini-Kinetics, it even improves the accuracy for all 3D CNNs. We will see that this is the case for all large datasets. For X3D-XL \cite{x3d}, we observe a $\sim$45\% reduction in GFLOPs and 0.2\% improvement in accuracy. We can see that {X3D-XL+ATFR}$^\beta$ requires similar GFLOPs compared to X3D-L$^\beta$ \cite{x3d} while providing a higher accuracy by 1.8\%. We can also see that {X3D-XL+ATFR}$^\alpha$ requires drastically less GFLOPs compared to X3D-L$^\beta$\cite{x3d} while getting a higher accuracy by 1.1\%.
In comparison to the computational heavy SlowFast16$\times$8,R101+NL \cite{slowfast}, {X3D-XL+ATFR}$^\beta$ gets higher top-5 and comparable top-1 accuracy while having 8.9$\times$ less GFLOPs. 

\begin{figure}[]
    \centering
    \includegraphics[width=1.0\columnwidth]{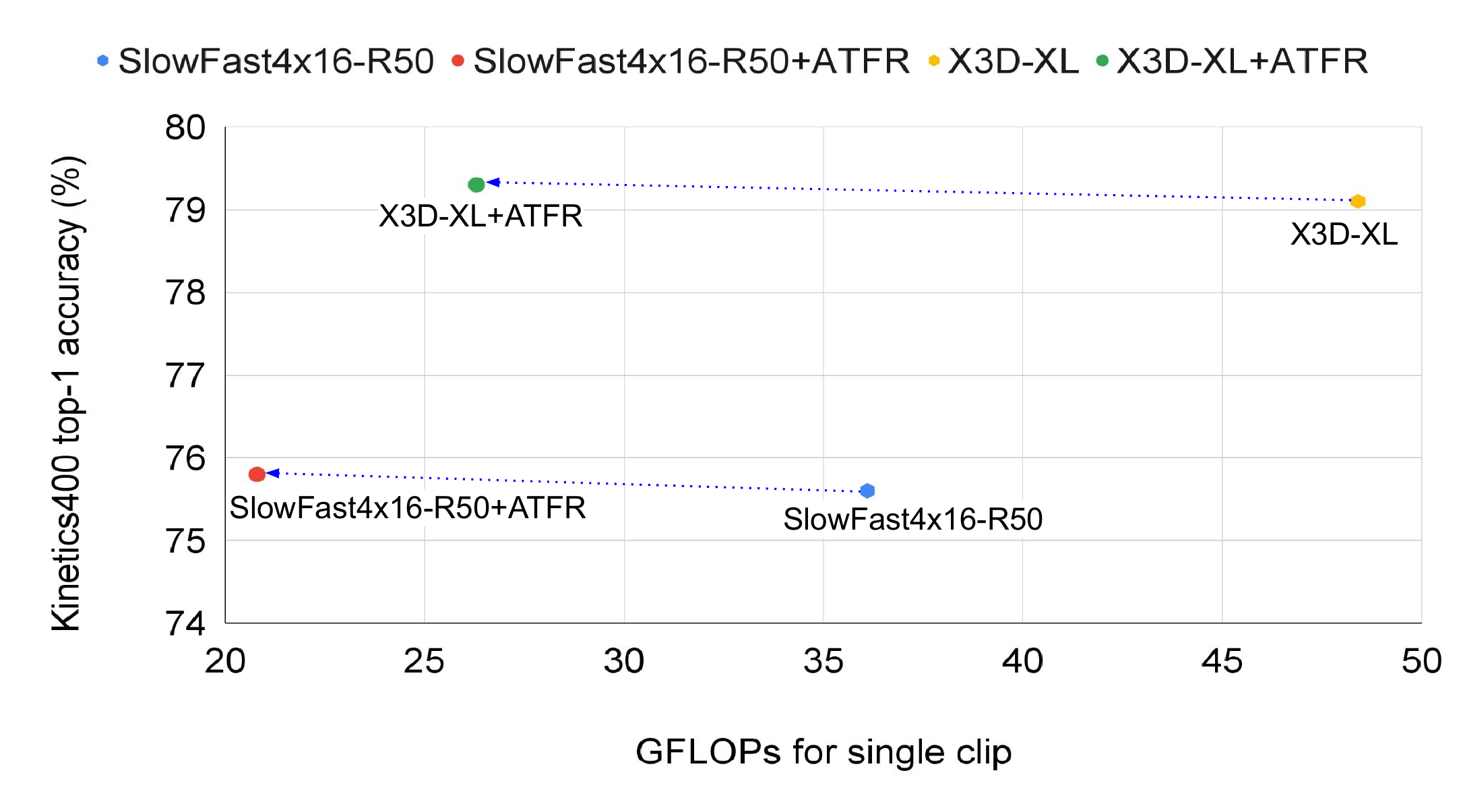}
    \caption{Accuracy vs.\ GFLOPs for the Kinetics-400 validation set.}
    \vspace{-8pt}
    \label{fig:accFlops}
\end{figure}

Comparing the 3D CNNs with ATFR to SCSampler \cite{SCSampler} and FASTER \cite{FASTER}, which require to train two networks, our approach with a single adaptive 3D CNN achieves a higher accuracy and lower GFLOPs. Note that our approach is complementary to \cite{SCSampler, FASTER} and the two static 3D CNNs used in these works can be replaced by adaptive 3D CNNs. Nevertheless, our approach outperforms these works already with a single 3D CNN. Fig.~\ref{fig:accFlops} shows the accuracy/GFLOPs trade-off for a few 3D CNNs with and without ATFR.


\begin{table}[t]
    \centering
    \resizebox{1.0\columnwidth}{!}{
    \begin{tabular}{|l|l|l|c|l|}
    \hline
        \multicolumn{1}{|c|}{model}   & \multicolumn{1}{c|}{GFLOPs} & \multicolumn{1}{|c|}{top1} & top5 & Param \\ \hline
        
        I3D$^*$\cite{i3d}                      & 108$\times$ N/A & 71.1 & 90.3 & 12.0M  \\ \hline
        
        I3D+SCSampler$^*$\cite{SCSampler}      &108$\times$10+N/A& 75.1 & N/A & N/A  \\ \hline
        
        Two-Stream I3D$^*$\cite{i3d}           &216$\times$N/A   & 75.7 & 92.0 & 25.0M  \\ \hline
        
        Two-Stream S3D-G$^*$\cite{miniKinetics1} &143$\times$N/A   & 77.2 & - & 23.1M \\ \hline
        
        TSM R50$^*$\cite{tsm}                 &65$\times$10     & 74.4 & N/A  & 24.3M \\ \hline \hline
        
        HATNET\cite{hatnet}               & N/A & 77.2 & N/A & N/A  \\ \hline
        
        STC\cite{stcnn}                   & N/A & 68.7 & 88.5 & N/A  \\ \hline
        
        Two-Stream I3D\cite{i3d}          &216$\times$N/A   & 75.7 & 92.0 & 25.0M  \\ \hline
        
        R(2+1)D\cite{R21}                 &152$\times$115     & 72.0 & 90.0 & 63.6M  \\ \hline
        
        Two-Stream R(2+1)D\cite{R21}      &304$\times$115     & 73.9 & 90.9 & 127.2M  \\ \hline
        
        FASTER32\cite{FASTER}             & 67.7$\times$8     & 75.3 & N/A & N/A
        \\ \hline
        

        SlowFast8$\times$8,R101+NL\cite{slowfast}& 116$\times$30  &78.7& 93.5 & 59.9M  \\ \hline
        
        SlowFast16$\times$8,R101+NL\cite{slowfast}& 234$\times$30  &\textbf{79.8}& 93.9 & 59.9M  \\ \hline
        
        X3D-L$^\alpha$\cite{x3d}          & 18.3$\times$10  & 76.8 & 92.5 & 6.1M \\ \hline
        
        X3D-L$^\beta$\cite{x3d}          & 24.8$\times$30  & 77.5 & 92.9 & 6.1M \\ \hline \hline
        
        SlowFast4$\times$16,R50\cite{slowfast}    & 36.1$\times$30 & 75.6        & 92.1 & 34.40M  \\ 
        
        SlowFast4$\times$16,R50\textbf{+ATFR}     & $20.8\times30 \  \textcolor{Blue}{(\downarrow42\%)} $& 75.8        & 92.4 & 34.40M \\ \hline
        
        
        X3D-S$^\alpha$\cite{x3d}          & 1.9$\times$10  & 72.9 & 90.5 & \textbf{3.79M} \\
        
        X3D-S\textbf{+ATFR}$^\alpha$      &$\textbf{1.0}\times10\ \ \ \textcolor{Blue}{(\downarrow47\%)} $& 73.5 & 91.2 & \textbf{3.79M} \\ \hline
        
        X3D-XL$^\alpha$\cite{x3d}        & 35.8$\times$10 & 78.4 & 93.6 & 11.09M  \\
        
        X3D-XL\textbf{+ATFR}$^\alpha$     & $20\times10\ \ \ \,  \textcolor{Blue}{(\downarrow44\%)} $& 78.6 & 93.9 & 11.09M \\ \hline
        
        X3D-XL$^\beta$\cite{x3d}          & 48.4$\times$30 & 79.1 & 93.9 & 11.09M \\ 
        
        X3D-XL\textbf{+ATFR}$^\beta$      & $26.3\times30 \, \textcolor{Blue}{(\downarrow45\%)} $& 79.3 & \textbf{94.1} & 11.09M \\ \hline
        \end{tabular}} \hfill
    \caption{Comparison to the state-of-the-art on Kinetics-400. X3D XL+AFTR$^{\beta}$ achieves the STA top5 while requiring 8.8$\times$ less GFLOPs compared to STA SlowFast16$\times$8,R101+NL. Following \cite{x3d}, we apply two testing strategies: $^\alpha$ samples uniformly 10 clips; $^\beta$ takes additionally 3 spatial crops for each sampled clip. For both setups, spatial scaling and cropping settings are as in \cite{x3d}. $^*$ denotes models pretrained on ImageNet.}
    \vspace{-8pt}
    \label{tab:Kinetics400}
\end{table}

\subsection{Kinetics-600}
We also evaluate our approach on the Kinetics-600 dataset \cite{kinetics600}. As shown in Table \ref{tab:Kinetics600}, ATFR shows a similar performance as on Kinetics-400. Our {SGS} module drastically decreases the GFLOPs of all 3D CNNs while improving their accuracy. For X3D-XL\cite{x3d}, we observe a $\sim$47.1\% reduction of GFLOPs and a slight improvement in accuracy. The best model {X3D-XL+ATFR} achieves state-of-the-art accuracy. Note that the average GFLOPs of {X3D-XL+ATFR} are even lower on Kinetics-600 compared to Kinetics-400. This shows that the additional videos of Kinetics-600 are less challenging in terms of motion, which is also reflected by the higher classification accuracy. Compared to SlowFast16$\times$8, R101+NL \cite{slowfast}, it requires about 9$\times$ less GFLOPs. 

\begin{table}[t]
    \centering
    \resizebox{1.0\columnwidth}{!}{
    \begin{tabular}{|l|c|l|c|c|}
    \hline
        \multicolumn{1}{|c|}{model}   & \multirow{1}{*}{pretrain}  & \multicolumn{1}{c|}{GFLOPs} & \multicolumn{1}{|c|}{top1} & top5 \\ \hline
        
        Oct-I3D+NL\cite{i3d}              & ImageNet & 25.6$\times$30 & 76.0 & N/A  \\ \hline
        
        HATNET\cite{hatnet}               & HVU      & N/A             & 81.6 & N/A  \\ \hline
        
        HATNET\cite{hatnet}               & -        & N/A             & 80.2 & N/A   \\ \hline
        
        I3D\cite{i3d}                     & -        & 108$\times$ N/A & 71.9 & 90.1  \\ \hline

        SlowFast16$\times$8,R101+NL\cite{slowfast}&-& 234$\times$30    &81.8  & 95.1  \\ \hline
        
        SlowFast4$\times$16,R50\cite{slowfast}    &-& 36.1$\times$30   &78.8  & 94.0  \\ \hline \hline
        
        X3D-M\cite{x3d}           & -     & 6.2$\times$30  & 78.8 & 94.5 \\
        
        X3D-M\textbf{+ATFR}       & -     & $\textbf{3.3}\times30  \ \ \,  \textcolor{Blue}{(\downarrow46\%)}  $& 79.0 & 94.9 \\ \hline
        
        X3D-XL\cite{x3d}          & -     & 48.4$\times$30  & 81.9 & 95.5 \\ 
        
        X3D-XL\textbf{+ATFR}      & -     & $25.6\times30  \  \textcolor{Blue}{(\downarrow47\%)} $& \textbf{82.1} & \textbf{95.6} \\ \hline
        \end{tabular}} \hfill
    \caption{Comparison to the state-of-the-art on Kinetics-600.}
    \label{tab:Kinetics600}
\end{table}

\subsection{Something-Something-V2}
We finally provide results for the Something-Something V2 dataset \cite{ssv}. It contains 169K training and 25K validation videos of 174 action classes that require more temporal modeling compared to Kinetics. Following \cite{multigrid}, we use a R50-SlowFast model pre-trained on Kinetics-400 with 64 frames for the fast pathway, speed ratio of $\alpha=4$, and channel ratio $\beta=1/8$. Similar to Kinetics, the SGS module reduces the GFLOPs by 33.9\% while keeping the accuracy almost the same. For more implementation details please refer to the supplementary material.
\begin{table}[t]
    \centering
    \resizebox{1.0\columnwidth}{!}{
    \begin{tabular}{|l|c|l|c|c|}
    \hline
        \multicolumn{1}{|c|}{model} & \multirow{1}{*}{pretrain} & \multicolumn{1}{c|}{GFLOPs} & \multicolumn{1}{c|}{top1}  & top5  \\ \hline 
        SlowFast-R50 \cite{multigrid}& Kinetics400 & 132.8 & 61.7 & 87.8    \\
        SlowFast-R50\textbf{+ATFR}  & Kinetics400 & \textbf{87.8}\ \ \,  \textcolor{Blue}{($\downarrow 33\%$)} & \textbf{61.8} & \textbf{87.9} \\ \hline
        \end{tabular}} \hfill
    \caption{Results for the Something-Something-V2 dataset. }
    \label{tab:ssv2}
\end{table}

\section{Conclusion}
Designing computationally efficient deep 3D convolutional neural networks for understanding videos is a challenging task. In this work, we proposed a novel trainable module called Similarity Guided Sampling (SGS) to increase the efficiency of 3D CNNs for action recognition. The new SGS module selects the most informative and distinctive temporal features within a network such that as much temporal features as needed but not more than necessary are used for each input clip.     
By integrating SGS as an additional layer within current 3D CNNs, which use static temporal feature resolutions, we can convert them into much more efficient 3D CNNs with \emph{adaptive temporal feature resolutions (ATFR)}. We evaluated our approach on six action recognition datasets and integrated SGS into five different state-of-the-art 3D CNNs. The results demonstrate that SGS drastically decreases the computation cost (GFLOPS) between 33\% and 53\% without compromising accuracy. For large datasets, the accuracy even increases and the 3D CNNs with ATFR are not only very efficient, but they also achieve state-of-the-art results.   



\paragraph{Acknowledgement} The work has been financially supported by the ERC Starting Grant ARCA (677650).

\newcount\cvprrulercount
\appendix
\section*{Appendix}

\setcounter{table}{0}
\renewcommand{\thetable}{A.\arabic{table}}
\renewcommand{\thefigure}{A.\arabic{figure}}

\section{Implementation Details}
\label{sup:imp_detailes}
\noindent\textbf{Modified 3DResNet-18} The architecture of our modified 3DResNet-18 is shown in Table \ref{table:hybridBackbone}. In case of 3DResNet-18+ATFR, we place SGS after the \textit{ResBlock 2}.

\begin{table}[h] 
\centering
\small
\tabcolsep=0.15cm

\begin{tabular}{ c| c| c }
  stage  & layer & output size \\
  \hline \hline 
  raw & - & 32 $\times$ 244 $\times$ 224  \\
  \hline
  conv$_{1}$ & 5 $\times$ 7 $\times$ 7, 8, stride 1, 2, 2 & 32 $\times$ 112 $\times$ 112 \\
  \hline
  pool$_{1}$ & $1 \times 3 \times 3$, max, stride $1, 2, 2$ & $32 \times 56 \times 56$ \\
  \hline
  res$_{2}$ &

    $
    \begin{aligned}
    \begin{bmatrix}
    3 \times 1 \times 1, 8 
    \\
    1 \times 3 \times 3, 8 
    \\
    1 \times 1 \times 1, 32
    \end{bmatrix} \times 2
    \end{aligned}
    $

& $32 \times 56 \times 56$ \\
  \hline
      res$_{3}$ & 
$
    \begin{bmatrix}
    \begin{aligned}
    1 \times 1 \times 1, 64 
    \\
    1 \times 3 \times 3, 64 
    \\
    1 \times 1 \times 1, 256
    \end{aligned}
    \end{bmatrix} \times 2
$
& $32 \times 28 \times 28$ \\
  \hline
      res$_{4}$ & 
$
    \begin{bmatrix}
    \begin{aligned}
    1 \times 1 \times 1, 128 
    \\
    1 \times 3 \times 3, 128 
    \\
    1 \times 1 \times 1, 512
    \end{aligned}
    \end{bmatrix} \times 2
$
& $32 \times 14 \times 14$ \\
  \hline
      res$_{5}$ & 
$
    \begin{bmatrix}
    \begin{aligned}
    3 \times 1 \times 1, 256 
    \\
    1 \times 3 \times 3, 256 
    \\
    1 \times 1 \times 1, 1024
    \end{aligned}
    \end{bmatrix} \times 2
$
& $32 \times 7 \times 7$ \\
  \hline
      & 
      global average pool, fc  
& $1 \times 1 \times 1$ \\
  \hline
   
\end{tabular}
\caption{ Modified 3DResNet-18 }
\label{table:hybridBackbone}
\end{table}

\begin{table}[]
\centering
\resizebox{0.9\columnwidth}{!}{
\begin{tabular}{ |l| c c |}
    \hline
      Model & Train & Inference (fps) \\
      \hline 
      X3D-S & 131h & 2834 \\
      X3D-S+ATFR & \textbf{121h} & \textbf{4295} \\
      \hline
    \end{tabular} \hfill
}
\caption{Runtime on Kinetics-400. 
}
\label{table:runtime}
\end{table}

\noindent\textbf{SlowFast-8x8-R50+ATFR}
Following \cite{multigrid} for training on the Something-Something-V2 dataset, the input temporal length to the SlowFast-8x8-R50+ATFR is set to 64. Due to the intensive size of the temporal domain, we limit the the temporal domain size of the SGS for each path-way. For the fast path-way we set the temporal domain size to 8. In other words, SGS is applied over temporal blocks with temporal length of 8 and temporal stride of 8. For the slow path-way we set the temporal domain size to 2. Since we drop zero bins in SGS, this may cause size mismatch for fusion in lateral connections. We therefore zero pad the smaller size tensors to the bigger ones.   

\section{Runtime}
To evaluate the runtime, we use X3D-S as the base model and report the runtimes for training and inference. As shown in Table~\ref{table:runtime}, SGS reduces the training time on Kinetics by $10$h. The ATFR equipped model processes almost $51\%$ more frames per second (fps) during inference. Our approach also requires less memory and we can use a larger batch size (BS), namely 256 instead of 208. 
This shows that the proposed approach substantially reduces GFLOPs, training and inference time, and memory usage. 

\section{Different number of bins}

The number of the sampling bins $B$ controls the maximum number of possible output feature maps of the SGS module. By setting $B=T$, the SGS module can keep all feature maps in case it is needed. To study the effect of changing $B$, we have evaluated the model performance by changing $B$ during training and inference. The base model is the 3DResNet-18 (Fig.~\ref{table:hybridBackbone}) trained on Mini-Kinetics. As it can be seen in Table \ref{table:numBinsTrain}, reducing $B$ decreases the accuracy, but also the GFLOPS. This is expected since SGS is forced to discard information for each video if $B < T$ even if there is no redundancy among the feature maps.      

As a second experiment, we change the number of bins only for inference while we train the model with $B=32$. This setting is interesting since it shows how flexible the approach is and if GFLOPS can be reduced at inference time without the need to retrain the model. The results are shown in Table \ref{table:numBinsInference}. If we compare the results with Table \ref{table:numBinsTrain}, we observe that the accuracy for training with $B=32$ and testing with $B=16$ is only slightly lower than training and testing with $B=16$. This shows that the GFLOPS can be reduced on the fly if it is required. However, if the difference between the number of bins during training and during inference is getting larger, the accuracy drops.

\begin{table}[]
    \centering
    \begin{tabular}{ |c| c c c c|}
        \hline
          B & 4 & 8 & 16 & 32 \\
          \hline 
          top1 & 61.4 & 64.7 & 64.7 & 69.6\\
          top5 & 86.3 & 85.8 & 86.2 & 88.8\\
          GFLOPs & 3.5 & 4.2 & 5.5 & 14.0  \\
          \hline
        \end{tabular}
    \caption{Ablations on the effect of changing the numbers of bins $B$ for 3DResNet-18+ATFR on Mini-Kinetics. The model is trained and validated for different number of bins. We show top-1 and top-5 classification accuracy (\%).}
    \label{table:numBinsTrain}
\end{table}

\begin{table}[]
    \centering
    \begin{tabular}{ |c| c c c c|}
        \hline
          B & 4 & 8 & 16 & 32 \\
          \hline 
          top1 & 51.1 & 61.4 & 64.4 & 69.6\\
          top5 & 75.1 & 83.5 & 85.5 & 88.8\\
          GFLOPs & 3.5 & 5.0 & 8.0 & 14.0 \\
          \hline
        \end{tabular}
    \caption{Ablations on the effect of changing the numbers of bins $B$ only during inference for 3DResNet-18+ATFR on Mini-Kinetics. The model is trained with 32 bins, but inference is performed with a different number of bins. We show top-1 and top-5 classification accuracy (\%). }
    \label{table:numBinsInference}
\end{table}

\section{Cartesian/Spherical Coordinates}
As mentioned in in Sec.~5.2.1, we use the magnitude of the embedding vectors as the similarity measurement to create the similarity bins. Instead of magnitudes, we can use other measures. While the results are reported in Table 1 of the paper, we describe how the approach works with spherical coordinates. 

 To use the spherical coordinates of the vectors for creating the similarity bins, we use multi-dimensional bins and sampling kernels. In an $L$ dimensional spherical coordinate system, we can use different subsets of coordinates for $\Delta_{t}^{k}$ with varying number of elements $K$ to create similarity bins, \eg, $K=L$ when using all of the coordinates, $K=L-1$ when using angular coordinates, or $K=1$ when using the radial coordinate only. Therefore, similar to Eq.~(2) and (3) of the paper, we can estimate $\beta_{b}^{k}$ for every $\Delta^{k}$.

By using a sampling kernel $\Psi(\Delta_{t}^{k}, \beta_{b}^{k})$ as in Eq.~(4) of the paper but for each $k$, a differentiable multi-dimensional sampling operation can be defined by
\begin{equation}
    \label{eq:GeneralMultiDimSamplingKernel}
    \mathcal{O}_{b}=\sum_{t=1}^{T} \mathcal{I}_{t} \prod_{k=1}^{K} \Psi(\Delta_{t}^{k}, \beta_{b}^{k}).
\end{equation}

\section{Similarity Guided Sampling Visualization}
The SGS layer aggregates similar input temporal feature maps into the same output feature map. To better understand such aggregation operation, we have visualized the input and output feature maps of the SGS layer in Figure \ref{fig:DTSN}. We have used a 3DResNet-50+ATFR trained on the Mini-Kinetics dataset. The sampling kernel used in this experiment is the linear kernel and the number of bins is set to 32. As it can be seen in Figure \ref{fig:DTSN}, the input temporal feature maps are aggregated to 4 output feature maps. The aggregated feature maps contain both the spatial and temporal information. In this example, the $4^{th}$ channel of the aggregated feature maps capture some motion flow that can be seen in the visualization.

\begin{figure*}[t]
    \centering
    \includegraphics[width=0.85\linewidth]{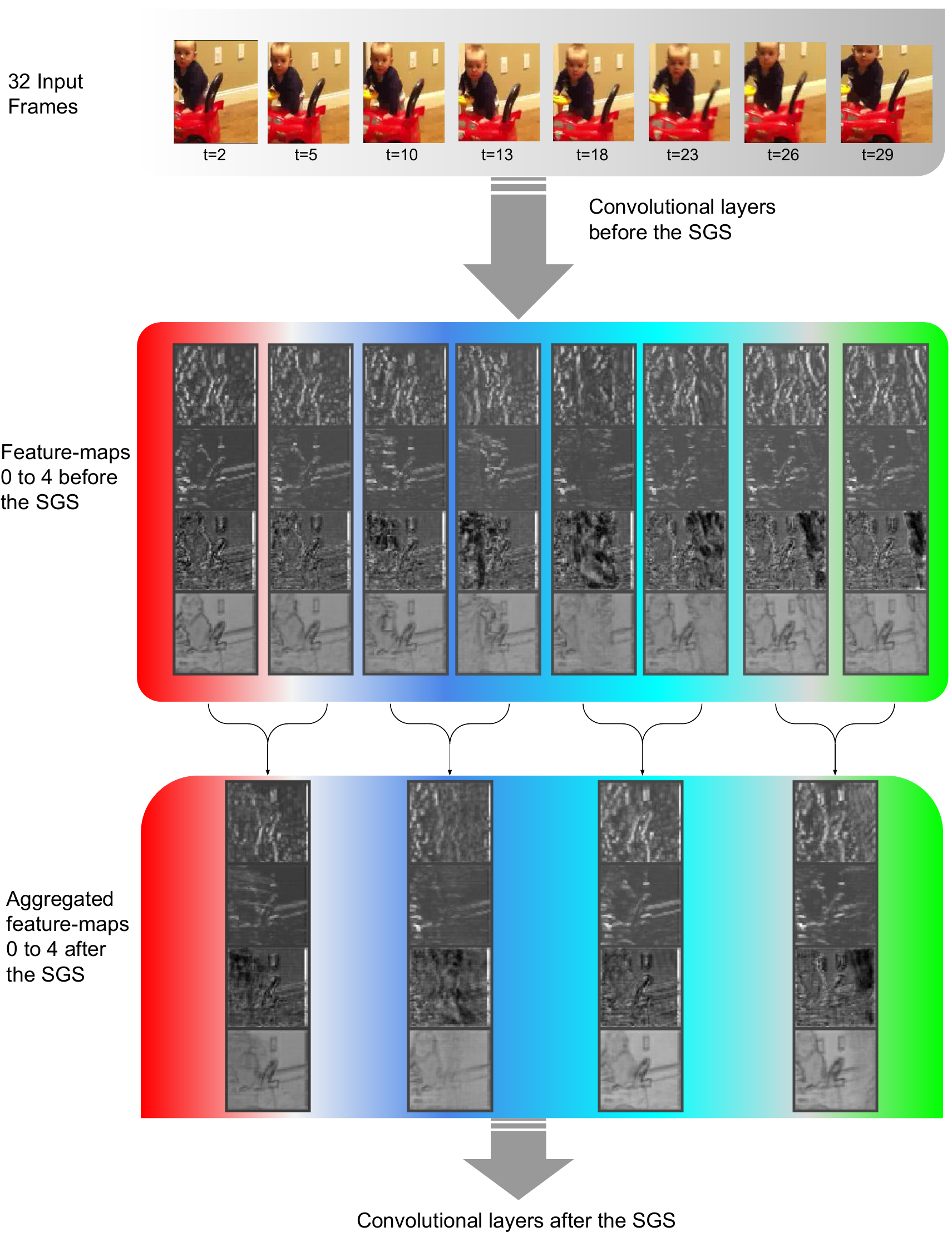}
    \caption{Visualization of the feature maps of 3DResNet-50+ATFR with linear kernel. In the first row, 8 frames out of 32 input frames are shown. The corresponding temporal feature maps of \textit{ResBlock 2} are depicted in the second row. The third row shows the aggregated feature maps after the SGS. Note that we only show the first 4 channels of the feature maps for better visualization.  }
    \label{fig:DTSN}
\end{figure*}

\section{UCF101 and HMDB51 Results}
\textbf{UCF-101} \cite{ucfdataset} contains 13K videos with 101 action classes. It is split into 3 splits with around 9.5K videos in each. For this dataset, we report the average accuracy over three splits.

\textbf{HMDB-51} \cite{hmdbdataset} has about 7000 videos with 51 action classes. It contains 3 splits for training and validation. Similar to UCF-101, we report the average accuracy over three splits. Table \ref{tab:ucfhmdb} shows the results on UCF-101 and HMDB51. The GFLOPs of our 3DResNet-R50+ATFR on UCF-101 and HMDB-51 are 22.2 and 23.1, respectively. As it can be seen, 3DResNet+ATFR gets comparable results compared to other 3D CNNs while having less GFLOPs as discussed in the paper.

\begin{table}[t]
    \centering
    \resizebox{1.0\columnwidth}{!}{
    \begin{tabular}{|l|c|c|c|}
    \hline
        \multicolumn{1}{|c|}{model} & backbone  & UCF101 & HMDB51 \\ \hline 
        C3D \cite{c3d} & RenNet18 & 89.8   & 62.1   \\ \hline
        RGB-I3D \cite{i3d} & Inception V1 & 95.6   & 74.8   \\ \hline
        R(2+1)D \cite{R21} & ResNet50 & 96.8   & 74.5   \\ \hline
        DynamoNet \cite{dynamonet} & ResNet101 &  96.6   & 74.9   \\ \hline
        HATNet \cite{hatnet} & ResNet50 &  97.8   & 76.5   \\ \hline 
        3DResNet+ATFR   & ResNet50 &  97.9   & 76.7  \\ \hline
    \end{tabular}}
    \caption{Comparison with other methods on UCF101 and HMDB51.}
    \label{tab:ucfhmdb}
\end{table}

\section{Comparison to Attention/Gating Mechanisms}
To better analyze the effect of our similarity guided sampling mechanism, we add attention modules to the base model and compare the final accuracy and GFLOPs to the base model and the ATFR model. To this end, we use a temporal attention mechanism following \cite{double_attn}. Similar to our SGS module, we add this attention module on top of the ResBlock2. As it can be seen in Table \ref {tab:attention}, the model equipped with the attention module achieves similar accuracy while requiring higher GFLOPs compared to the model equipped with SGS. The reason for such a great difference in GFLOPs is that attention modules perform a weighting of the features, while our approach clusters and reduces features. If all features are the same, the attention module should weight them equally while our approach reduces them to one feature.

\begin{table}[t]
    \centering
    \resizebox{1.0\columnwidth}{!}{
    \begin{tabular}{ |l| c c |}
    \hline
      Model & top1 & GFLOPs \\
      \hline 
      X3D-S & 77.9 & 1.9 \\
      X3D-S+ATFR & 78.0 & \textbf{1.1} \\
      X3D-S+Temporal Attention & 78.3 &  1.9 \\
      \hline
    \end{tabular} \hfill}
    \caption{Comparison with attention modules. The models are trained and tested on the Mini-Kinetics dataset.}
    \label{tab:attention}
\end{table}

{\small
\bibliographystyle{ieee_fullname}
\bibliography{references}
}

\end{document}